\documentclass[journal,onecolumn]{IEEEtran}

\usepackage{amsmath,amsfonts}
\usepackage{algorithmic}
\usepackage{algorithm}
\usepackage{array}
\usepackage[caption=false,font=normalsize,labelfont=sf,textfont=sf]{subfig}
\usepackage{textcomp}
\usepackage{stfloats}
\usepackage{url}
\usepackage{verbatim}
\usepackage{graphicx}
\usepackage{cite}
\usepackage[colorlinks,
            linkcolor=blue,
            anchorcolor=blue,
            citecolor=blue]{hyperref}
\usepackage[T1]{fontenc}

\usepackage{subfig}

\usepackage{booktabs}
\usepackage{multirow}
\usepackage{makecell}

\begin{document}

\title{Bidirectional Temporal Information Propagation for Moving Infrared Small Target Detection}

\author{Dengyan~Luo, Yanping~Xiang,
Hu~Wang, Luping~Ji, ~\IEEEmembership{Member,~IEEE},  Shuai~Li, ~\IEEEmembership{Senior Member,~IEEE,} and Mao~Ye$^*$, ~\IEEEmembership{Senior Member,~IEEE}

\thanks{This work was supported in part by the National Natural Science Foundation of China (62276048) and Chengdu Science and Technology Projects (2023-YF06-00009-HZ).}
\thanks{Dengyan Luo, Yanping Xiang, Hu Wang, Luping Ji and Mao Ye are with the School of Computer Science and Engineering, University of Electronic Science and
Technology of China, Chengdu 611731, P.R. China (e-mail: dengyanluo@126.com;
xiangyp@uestc.edu.cn;  wanghu0833cv@gmail.com; jiluping@uestc.edu.cn;
cvlab.uestc@gmail.com).}
\thanks{Shuai Li is with the School of Control Science and Engineering, Shandong University, Jinan 250000, P.R. China (e-mail:shuaili@sdu.edu.cn).}
\thanks{*corresponding author}}


\markboth{}%
{Shell \MakeLowercase{\textit{et al.}}: A Sample Article Using IEEEtran.cls for IEEE Journals}



\maketitle

\begin{abstract}
Moving infrared small target detection is broadly adopted in infrared search and track systems, and has attracted considerable research focus in recent years.
The existing learning-based multi-frame methods mainly aggregate the information of adjacent frames in a sliding window fashion
to assist the detection of the current frame. 
However, the sliding-window-based methods do not consider joint optimization of the entire video clip and ignore the global temporal information outside the sliding window,
resulting in redundant computation and sub-optimal performance.
In this paper, we propose a Bidirectional temporal information propagation method for
moving InfraRed small target Detection, dubbed BIRD. 
The bidirectional propagation strategy simultaneously utilizes local temporal information of adjacent frames and global temporal information of past and future frames in a recursive fashion.
Specifically, in the forward and backward propagation branches, 
we first design a Local Temporal Motion Fusion (LTMF) module to model local spatio-temporal dependency between a target frame and its two adjacent frames.
Then, a Global Temporal Motion Fusion (GTMF) module is developed to further aggregate the global propagation feature with the local fusion feature.
Finally, the bidirectional aggregated features are fused and input into the detection head for detection.
In addition, the entire video clip is jointly optimized by the traditional detection loss and the additional Spatio-Temporal Fusion (STF) loss.
Extensive experiments demonstrate that the proposed  BIRD method not only 
achieves the state-of-the-art performance but also shows a fast inference speed.
\end{abstract}


\section{Introduction}
\label{Sec1}
Moving Infrared Small Target Detection (MISTD) technology 
is broadly adopted in many real applications,
such as military and industrial surveillance, because 
infrared images can highlight targets in any illuminance condition \cite{peng2023courtnet}.
Although small object detection and video target detection   technologies \cite{xu2021video,9747993,10173615,wang2024tiny,chen2024dila}
have been developed rapidly
over the past decades,
MISTD task still faces huge challenges.
Firstly, due to the long imaging distance and low resolution of infrared imaging systems, 
infrared targets are typically   small  and lack texture information \cite{10298041,feng2024meta}, 
which can lead to missed detections and false detections. 
Secondly, a significant amount of noise and a cluttered background can easily disturb or obscure infrared small targets.
Thirdly, the fast motion of targets 
often causes the  boundary area of targets to suffer severe distortions.
Therefore,  improving the detection performance of MISTD is a very meaningful  and challenging task.

Over the past few decades, many works have been proposed for MISTD task.
The  existing works 
can be categorized into
{\it traditional} and {\it learning-based} methods.
Although conventional methods, for instance, 
optical flow-based method \cite{kwan2020enhancing,zhao2019infrared}
and tensor optimization-based method \cite{sun2019infrared,wu2023infrared}, 
can improve detection performance to a certain degree, 
they are hindered by extensive hand-crafted components.
Once the scene becomes too complex, they tend to produce false alarms and missed detections.
Different from traditional methods,
learning-based methods can utilize deep neural networks to learn features from numerous samples and 
show strong flexibility and capability in complex scenes.
Therefore, learning-based methods have become the mainstream scheme in recent years.

According to the difference in the number of frames used, 
the existing learning-based methods can be further divided into two categories: {\it single-frame} and {\it multi-frame} approaches.
Single-frame methods \cite{chen2024tci,LIN2024124385,chen2024unveiling} 
detect targets in an image at a time.
Although these single-frame methods can
detect objects in video in a frame-by-frame manner,
they are less effective because they fail to exploit valuable temporal information, 
which can be particularly beneficial for identifying objects whose boundaries are blurred due to fast motion.
Thus, the multi-frame approach has received significant attention in the past few years.

\begin{figure}[t]
  \centering
  \includegraphics[width=0.55\linewidth]{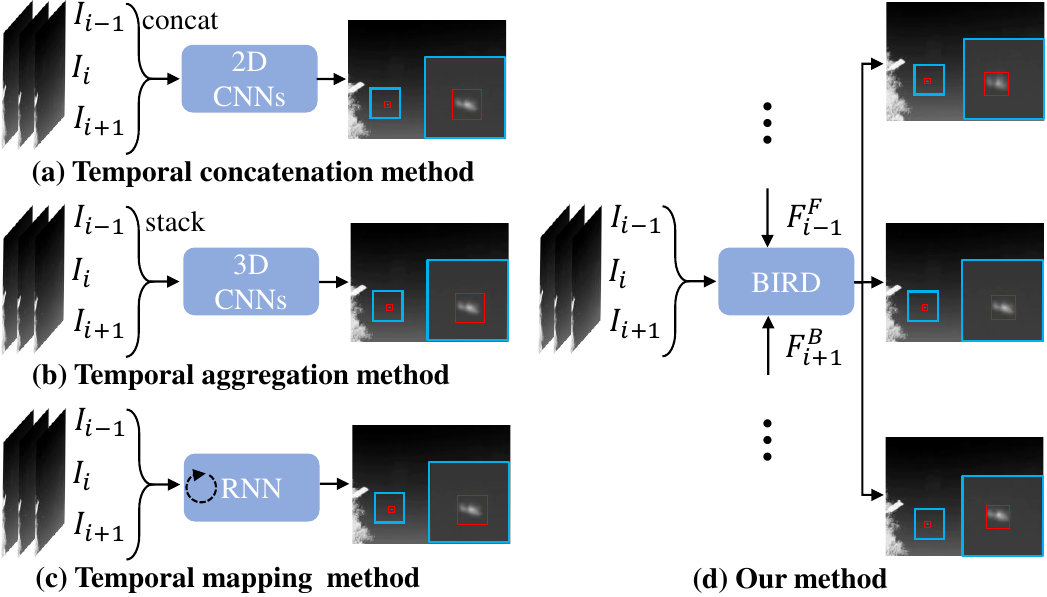}
  \caption{
  Illustrating the differences between our method and the existing learning-based multi-frame methods.
  The existing three kinds of methods (a-c) process the video sequence in a sliding window fashion (i.e., inputting one video clip and outputting the detection result of one frame at a time).
Our method (d) processes the video sequence in a recursive fashion, by inputting one video clip and outputting the detection results of the entire video clip at a time. 
It can aggregate global temporal information from distant frames, unlike the existing methods that only aggregate local temporal information from neighboring frames.  The $F_{i-1}^F$ and $F_{i+1}^B$ represent the distant past (forward) and future (backward) propagation  features of the current sampling window, respectively.
  }
  \label{compare}
\end{figure}

According to the way of leveraging inter-frame temporal information, 
the existing learning-based multi-frame based approaches can be roughly divided into three types:
1) temporal concatenation \cite{du2021multiple,du2021spatial,yan2023stdmanet,10521471,ZHU2024124731}, 2) temporal aggregation \cite{10275009,10321723,tong2024st} and 3) temporal mapping \cite{liu2021dim,chen2024sstnet}, 
as shown in Fig. \ref{compare} (a), (b) and (c), respectively.
The temporal concatenation scheme directly concatenates multiple frames or features in the channel axis, and uses a 2D Convolutional Neural Network (CNN) to fuse the temporal information.
The temporal aggregation scheme applies a 3D CNN to extract both spatial and temporal information in the video clip. 
The temporal mapping scheme maps multi-frame features to sentence and modeling motion information via a Recurrent Neural Network (RNN).

\begin{figure}[t]
  \centering
  \includegraphics[width=0.55\linewidth]{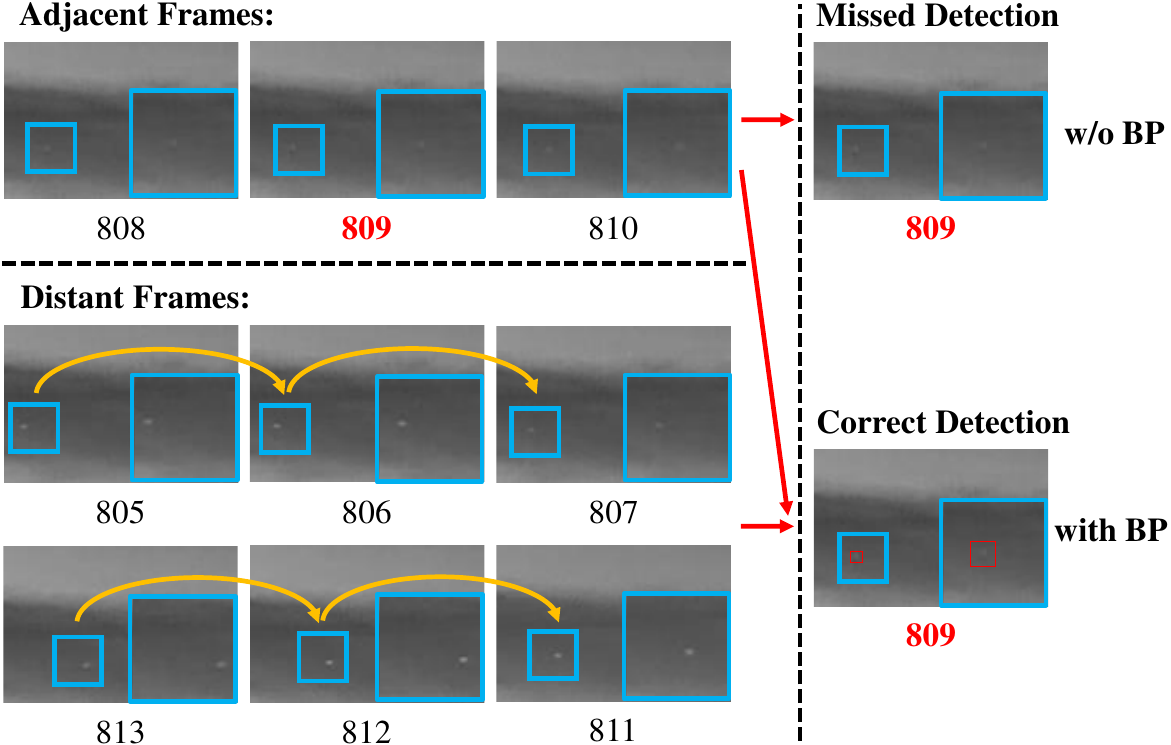}
  \caption{
  An example of using distant frame information. To detect the target in frame 809, only using local adjacent frames does not provide enough information to help the detection of the current frame because the targets are also difficult to perceive in these adjacent frames. The proposed Bidirectional Propagation (BP) scheme can utilize the corresponding  areas in distant frames to assist target detection in the current frame.
  }
  \label{why_BP}
\end{figure}

Although the above mentioned methods utilize both intra-frame spatial information and inter-frame temporal information, 
they process the video sequence in a sliding window fashion (i.e., dividing the video sequence into multiple clips, and inputting one video clip and outputting the detection result of one frame at a time), 
so they have the following obvious disadvantages:
(1) They fail to consider the joint optimization of the entire video clip in the training stage, resulting in sub-optimal performance.
(2) Due to the many-to-one input-output mapping, 
when processing the whole video sequence  in the inference stage, 
each video frame is processed multiple times, 
resulting in heavy computational overhead and expensive time costs.
(3) 
In the inference stage, they can only utilize the local temporal information of adjacent frames within a narrow sampling window, 
while ignoring the global temporal information of distant frames outside the sliding window.
The essence is that once the models for these methods are trained, the number of frames used in their inference and training  stages must remain consistent.
However, the global distant frame information is very important. 
As shown in Fig. \ref{why_BP}, 
the targets in the current frame 809  and its adjacent frames 808 and 810 are difficult to detect, 
while the targets in the distant frames 805, 806 and 811-813 are easy to perceive.

If many-to-many input-output mapping can be achieved, 
the above mentioned limitations (1) and (2) can be effectively alleviated. If the temporal information outside the sampling window can be utilized,
the limitation (3) can be mitigated.
Therefore, in this paper,
we propose a Bidirectional temporal information propagation method for
moving InfraRed small target Detection, called BIRD. BIRD is based on a newly developed fashion, named recursive fashion, which looks around to utilize the entire video clip and propagate global temporal information, as illustrated in Fig. \ref{compare} (d).
Intuitively, there are two challenges that need to be addressed, i.e., the utilization of local neighboring frame information and the global distant frame information.

Specifically, for utilizing local neighboring frame information, a Local Temporal Motion Fusion (LTMF) module based on deformable convolution is designed to model both spatial correlation in the target frame and temporal dependence between consecutive frames. The LTMF module takes a target frame feature and its two adjacent frame features as input to jointly predict deformable convolution offsets. 
As a result, complementary information from the target and neighboring frames can be fused in an adaptive manner.
Furthermore, to stabilize the fusion of consecutive frames, we introduce a Spatio-Temporal Fusion (STF) loss $\mathcal{L}_{STF}$ by measuring the $L_1$ distance between the fused spatio-temporal feature and the target frame feature.
For aggregating global distant frame information, we propose a bidirectional propagation scheme to aggregate information from both past and future frames.
In the forward and backward propagation branches, a Global Temporal Motion Fusion (GTMF) module is developed to aggregate the local fusion feature and the global propagation feature. In each propagation branch, local and global temporal information can be utilized.
Finally, the bidirectional aggregated features are fused and input into the target detection head.
As we can see from Fig. \ref{why_BP}, when the bidirectional propagation scheme is not used, 
we cannot detect the target in frame 809, 
while our method with the bidirectional propagation scheme can successfully detect the target by exploiting the corresponding areas in the distant frames.

The primary contributions of this paper are summarized as follows:
(1) A novel BIRD method is proposed 
to effectively use motion context by a bidirectional propagation scheme. The proposed BIRD method processes the video sequence in a recursive fashion rather than a sliding window fashion.
(2) Local and global information are sufficiently used. A LTMF module is designed to model both spatial dependence of intra-frame and local temporal dependence of inter-frame. Additionally, a new STF loss $\mathcal{L}_{STF}$ is introduced to improve the fusion of consecutive frames. A GTMF module is further developed to aggregate the local fusion feature and the global propagation feature.
(3) Experimental results demonstrate that the proposed BIRD method can perform favorably against the state-of-the-art methods both quantitatively and qualitatively with faster inference speed.

\section{Related Works}
\label{Sec2}
\subsection{Single-frame infrared small target detection}
The single-frame ISTD  scheme processes an infrared image at a time and  can be categorized into traditional and learning-based approaches.

The traditional single-frame methods \cite{zhu2020tnlrs,gao2013infrared,liu2023combining,deshpande1999max,bai2010analysis,chen2013local,moradi2020fast} can be mainly divided into data structure-based, background modeling-based and target feature-based approaches. 
The data structure-based method \cite{zhu2020tnlrs,gao2013infrared,liu2023combining}
separates background and  targets in an infrared image based on the low rank of background and the sparsity of targets. 
The background modeling-based method  \cite{deshpande1999max,bai2010analysis}
estimates and mitigates the background effect  by predicting the background intensity value of each pixel. 
The target feature-based method  \cite{chen2013local,moradi2020fast}
emphasizes targets by utilizing 
the feature differences between targets and their contiguous regions.
Although these methods can improve detection performance
to a certain extent, they have extensive hand-crafted components (e.g., feature extractor and detector), resulting in poor detection performance in intricate scenes.

The learning-based single-frame methods 
\cite{wang2019miss,dai2021asymmetric,wang2022interior,zhang2022rkformer,10295542,sun2023receptive,10288394}
use deep neural networks to learn target features from  plenty of training samples. 
For example, Dai et al. \cite{dai2021asymmetric} designed a ACM to better highlight small targets by 
fusing low-level and deep-level features extracted by CNN.
Later on, 
to model long-range dependencies between pixels,
Wang et al. \cite{wang2022interior}
 developed a IAANet based on transformer for ISTD task.
After that, to take advantage of both CNN and transformer,
Zhang et al. \cite{zhang2022rkformer} developed a RKformer and 
Lin et al. \cite{10295542} proposed  a IR-TransDet  to extract local and global spatial features to improve detection performance.
Although these methods can also be used to address 
MISTD task, they are not very effective 
because they
ignore the valuable temporal information.

\subsection{Multi-frame infrared small target detection}
The multi-frame methods utilize the spatio-temporal information of consecutive multiple frames to detect targets in an infrared image.
These methods can also be divided into traditional  and learning-based  approaches.

The traditional multi-frame methods \cite{kwan2020enhancing,zhao2019infrared,sun2019infrared,wu2023infrared,zhang2005detecting,ren2019infrared}
can be mainly divided into 
optical flow-based, tensor optimization-based and 
energy accumulation-based  approaches.
The optical flow-based method \cite{kwan2020enhancing,zhao2019infrared} uses 
brightness changes in successive frames to represent target motion.
The tensor optimization-based method \cite{sun2019infrared,wu2023infrared} constructs a 4-D spatio-temporal tensor to accentuate targets.
The energy accumulation-based method \cite{zhang2005detecting,ren2019infrared} is based on the direction of motion to accumulate the energy of small targets.
Even though these methods have shown some advancement,
they also produce many false detections when the background is too complex.

To alleviate the above problems, learning-based multi-frame detection methods 
\cite{liu2021dim,du2021multiple,du2021spatial,10275009,yan2023stdmanet,10321723,10521471,tong2024st,chen2024sstnet} 
are proposed.
These methods can be further divided into three types: 1) temporal concatenation \cite{du2021multiple,du2021spatial,yan2023stdmanet,10521471,ZHU2024124731}, 2) temporal aggregation \cite{10275009,10321723,tong2024st} and 3) temporal mapping \cite{liu2021dim,chen2024sstnet}.
For the temporal concatenation approach, 
for example, Du et al. \cite{du2021spatial} developed a STFBD 
to extract spatial-temporal features through the inter-frame energy accumulation enhancement mechanism.
Afterwards, Zhu et al. \cite{ZHU2024124731} designed a TMP with two parallel branches to extract spatial and temporal features, respectively.
However, the simple direct concatenation fusion strategy does not consider the inter-frame alignment and thus may introduce new artifacts that degrade the detection performance.
For the temporal aggregation approach, 
for instance, 
Li et al. \cite{10321723} proposed a DTUM to 
transform temporal information into  features and extract the spatio-temporal feature by applying 3D CNN. 
Meanwhile, Tong et al. \cite{tong2024st}  utilized 3D CNN and transformer and proposed a ST-Trans for the detection of infrared  small targets in  sequential images.
Although 3D CNN performs implicit inter-frame alignment, 
it significantly increases the computational complexity and memory, which slows down the inference speed.
For the temporal mapping approach, 
Liu et al. \cite{liu2021dim} utilized 3D CNN and bidirectional ConvLSTM to 
model temporal motion.
Recently, Chen et al. \cite{chen2024sstnet} developed a SSTNet based on ConvLSTM and a motion-coordination loss to capture motion features.
More recently, Luo et al. \cite{LUO2026111894} proposed a DFAR based on deformable convolution to model temporal information.
Although temporal mapping approach has achieved the state-of-the-art performance,
LSTM-based models typically have a plethora of parameters and require considerable computational resources.
Furthermore, the SSTNet method utilized ConvLSTM  only  in the training stage, which may  result in overfitting.

Last but the most important is that 
all existing learning-based  multi-frame detection methods process the video
sequence in a sliding window fashion, 
resulting in individual frame optimization and the failure to utilize global temporal information.
Processing each input video frame more than once
also leads to redundant computation.
In contrast, 
our method processes the video sequence in a recursive fashion,
thus improving detection performance and speeding up inference speed.

\section{The Proposed Method}

\label{Sec3}
\subsection{Overview}
\label{sec3-1}
Given an infrared video $X = \left\{I_1, I_2, \ldots, I_T\right\}$ that is
composed of $T$ consecutive frames.
The task of MISTD is to detect targets in each video frame.
As discussed in  Sections \ref{Sec1} and \ref{Sec2}, the existing learning-based multi-frame methods take  consecutive multiple frames as input to get the detection result of one frame, 
and then process the entire video in a sliding window fashion to obtain the detection results of all frames.
On the contrary, theoretically, our method can use the entire video as input to simultaneously obtain detection results for all frames. 
However, due to limited memory, the entire video needs to be divided into multiple video clips for processing.
In fact, in our method, we take a video clip consisting of 
$N$ 
consecutive frames as input to get the detection results of all frames at once.
These input frames are denoted by
$I_{[t,t+N]}=\left\{I_{t}, \ldots, I_{t+N}\right\}$, 
where $I_{[t,t+N]} \in \mathbb{R}^{N \times C \times H \times W}$. 
Here, $C$ refers to the number of channels of a single frame, 
and $H \times W$ denotes the frame size.

\begin{figure*}[t]
    \centering
    \includegraphics[width=0.95\textwidth]{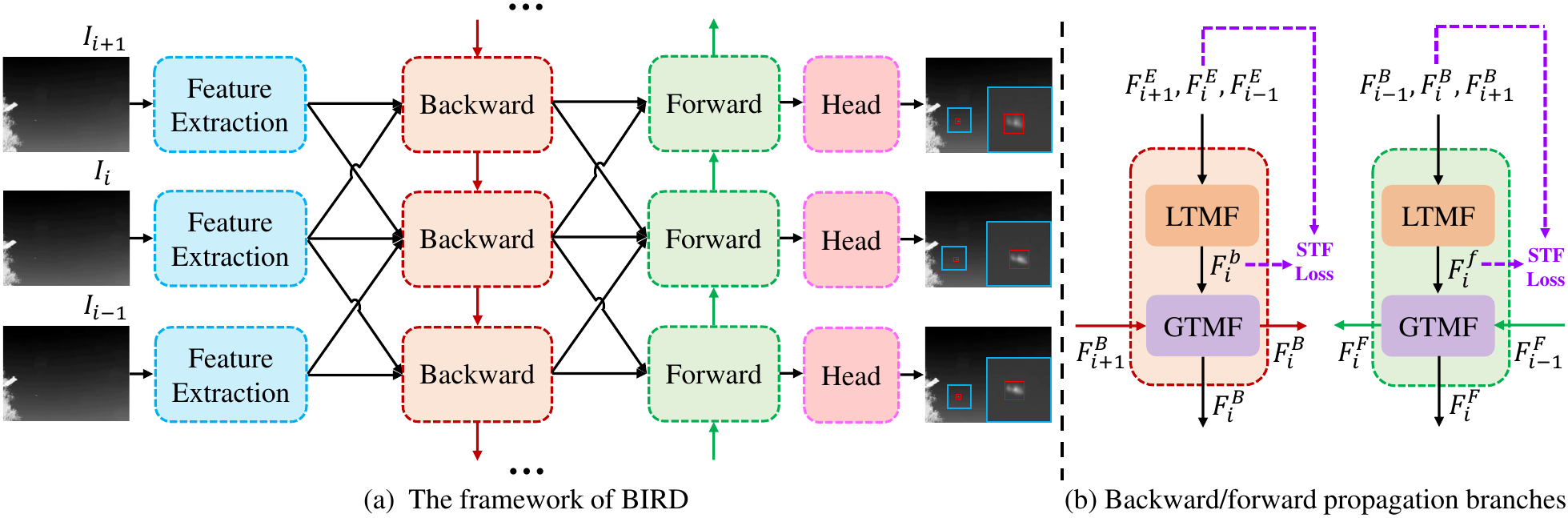}
\caption{\textbf{The overview of BIRD.} \textbf{(a)} 
(1) The feature extraction module is used to extract the spatial information of each frame.
(2) The extracted visual features and past backward propagation feature are fed into the backward propagation branch for aggregating future frame information.
(3) The backward propagation features and past forward propagation feature are input into the  forward propagation branch for aggregating past frame information.
(4) The fused bidirectional propagation features  are fed into the detection head module for calculating the detection loss.
\textbf{(b)} The propagation branches are based on Local Temporal Motion Fusion (LTMF) and Global Temporal Motion Fusion (GTMF) modules.
  The network is 
   is trained with
   the standard detection loss  and the introduced Spatio-Temporal Fusion (STF) loss.
 }
  \label{framework}
\end{figure*}

The BIRD framework is shown in Fig. \ref{framework} (a).
we first use the feature extraction module to 
extract the spatial information of each frame and  get the extracted features $F_{[t,t+N]}^{E} \in \mathbb{R}^{N \times c  \times h \times w}$, where $c$, $h$, and $w$ denote the channel, height, and width
of the feature $F_{i}^{E}$, respectively.
This process can be expressed as:
\begin{equation}
  F_{i}^{E}=FE(I_{i}), i \in [t,t+N],
\end{equation}
where $FE(\cdot)$ represents the feature extraction module, which is a three-level pyramid structure, with each layer consisting of a convolutional layer with a stride of 1 and another with a stride of 2.
Then, a backward propagation branch is used to aggregate local  temporal information and future frame information: 
\begin{equation}
  F_{i}^B = BP(F_{i+1}^E, F_i^E, F_{i-1}^E, F_{i+1}^B),   
\label{eq2}
\end{equation}
where $BP(\cdot) $ represents the backward propagation branch, 
$F_{i+1}^B$ is the past propagation feature with respect to the $i$-th frame,
which carries future frame information;
and $F_{i}^B \in \mathbb{R}^{c  \times h \times w}$ refers to the generated backward propagation feature of the $i$-th frame.
Furthermore, a forward propagation branch is employed
to aggregate past frame information:
\begin{equation}
  F_{i}^F = FP(F_{i-1}^B, F_i^B, F_{i+1}^B, F_{i-1}^F),   
\label{eq3}
\end{equation}
where $FP(\cdot) $ represents the  forward propagation branch, 
$F_{i-1}^F$ is also the past propagation feature with respect to the $i$-th frame, 
which carries past frame information; 
and $F_{i}^F \in \mathbb{R}^{c  \times h \times w}$ refers to the generated forward propagation feature of the $i$-th frame.
After that, a standard $3 \times 3$ convolutional layer is utilized to fuse extracted features and bidirectional propagation features:
\begin{equation}
  F_i^D = Conv([F_i^E, F_i^B, F_i^F]), 
\end{equation}
where $[\cdot,\cdot]$ denotes the concatenation operation in the channel axis, and $F_i^D$ is the fused spatio-temporal feature of the $i$-frame.
Finally, following SSTNet \cite{chen2024sstnet}, each fused feature $F_i^D$ is fed into the detection head YOLOX \cite{ge2021yolox} for target detection.

It should be noted that for a segmented video clip that has less than $N$ frames, 
we fill it with the last frame until there are $N$ frames.
Meanwhile, we initialize both propagation features $F_{t+N+1}^B$ and $F_{t-1}^F$ to 0.
The details of the backward and forward propagation branches are explained in the following Section \ref{3.2},
while the loss function is introduced in Section \ref{3.3}.

\subsection{Bidirectional propagation branches}
\label{3.2}
The existing
learning-based multi-frame methods
fail to utilize global temporal
information.
To simultaneously utilize local neighboring frame information and global distant frame information,
we design a bidirectional propagation scheme to maximize information gathering.
The structures of the backward and forward propagation branches are shown in Fig. \ref{framework} (b).
For the frame $I_i$,
in the backward propagation branch,   
first,
the Local Temporal Motion Fusion (LTMF) module takes the extracted feature $F_i^E$ and its adjacent frame features $F_{i+1}^E$ and $F_{i-1}^E$ as inputs and  produces the local fused feature $F_i^b$:
\begin{equation}
F_i^b = BP_{LTMF}(F_{i+1}^E, F_i^E, F_{i-1}^E), 
\end{equation}
where $BP_{LTMF}(\cdot) $ denotes the LTMF module in the backward propagation branch. 
Then, the Global Temporal Motion Fusion (GTMF) module uses the local fused feature $F_i^b$ and the past propagation feature $F_{i+1}^B$ as input and generates the new propagation feature $F_{i}^B$:
\begin{equation}
F_{i}^B = BP_{GTMF}(F_i^b, F_{i+1}^B), 
\label{eq6}
\end{equation}
where $BP_{GTMF}(\cdot) $ denotes the GTMF module in the backward propagation branch. Here, the output of Eq. (\ref{eq6}) corresponds to Eq. (\ref{eq2}).
The new propagation feature $F_{i}^B$ is not only input into the forward propagation branch, but also utilized by the next frame in the backward propagation branch.

Similarly, the forward propagation branch uses backward propagation features $F_{i-1}^B, F_i^B, F_{i+1}^B$ and past propagation feature $F_{i-1}^F$ to generate new propagation feature $F_{i}^F$:
\begin{eqnarray}
&&F_i^f  = FP_{LTMF}(F_{i-1}^B, F_i^B, F_{i+1}^B), \\
&&F_{i}^F  = FP_{GTMF}(F_i^f, F_{i-1}^F), 
\label{eq8}
\end{eqnarray}
where $F_i^f$ is the local fused feature in the forward propagation branch; $FP_{LTMF}(\cdot) $ and $FP_{GTMF}(\cdot) $ represent the LTMF module and GTMF module in the forward propagation branch, respectively. The output of Eq. (\ref{eq8}) corresponds to Eq. (\ref{eq3}).

It should be noted that in the above forward and backward propagation processes, 
there are not enough adjacent frames available at the boundary of clip, 
so we directly copy the features of the boundary frames for using. 
The modules in the forward and backward propagation branches have the same structure but do not share parameters.   
In the following subsections, we  only introduce the modules in the backward propagation branch, 
and the procedure for the modules in the forward propagation branch is similarly.

\subsubsection{Local temporal motion fusion  module}
The existing learning-based multi-frame methods aggregate locally adjacent frame information by direct concatenation or 3D CNN or LSTM. 
However, they are not sufficient to handle fast motion, 
which introduces new artifacts during the aggregation process, 
resulting in the generated target bounding boxes being inaccurate and too large.
Fortunately, deformable convolution (DCN)  \cite{dai2017deformable} 
has shown excellent 
performance at
modeling geometric transformations and aggregating information from distant spatial locations in some vision tasks, such as video frame interpolation \cite{9328265} and video quality enhancement \cite{luo2023spatio}.
Therefore, we design a LTMF module based on DCN to aggregate local temporal information.

\begin{figure*}[t]
  \centering
  \includegraphics[width=0.45\linewidth]{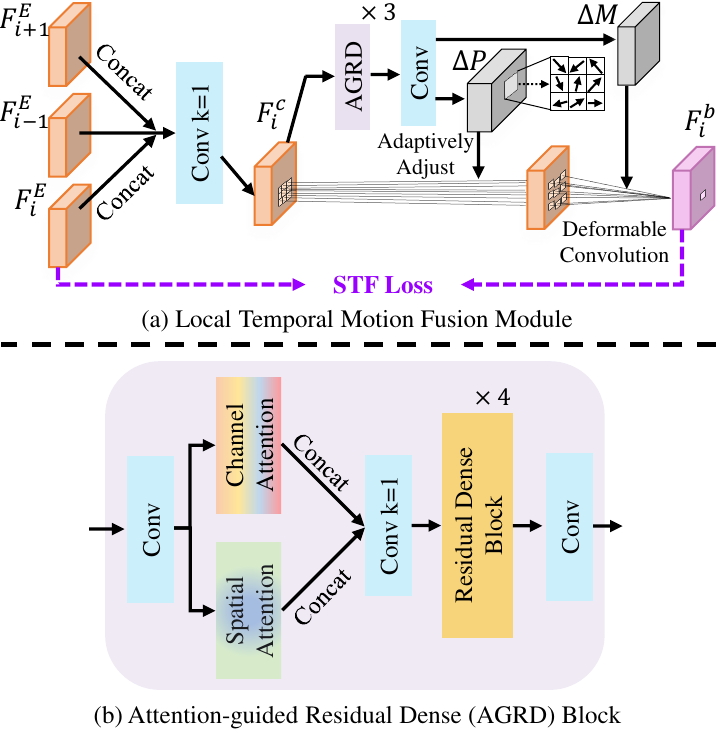}
  \caption{
  The details of Local Temporal Motion Fusion (LTMF) module in backward propagation branch. 
  With no special indication, 
  the kernel size of the convolutional layer is set to $3 \times 3$.
  }
  \label{LTMF}
\end{figure*}

Technically, as shown in Fig.  \ref{LTMF} (a), 
we first use a $1 \times 1$ bottleneck convolution to fuse the reference frame and its two adjacent frames: 
\begin{equation}
F_{i}^c = Conv([F_{i+1}^E, F_i^E, F_{i-1}^E]), 
\end{equation}
where $F_{i}^c$ is the concatenated feature.
Then, the stacked Attention-guided Residual Dense (AGRD) blocks and a $3 \times 3$ convolutional layer are used to generate the corresponding deformable sampling parameters:
\begin{equation}
 \Delta P_i, \Delta M_i = Conv \circ (AGRD)^3(F_{i}^c),
\end{equation}
where $ \Delta P \in \mathbb{R}^{(d \times 2 K^{2}) \times h \times w}$  and  $ \Delta M  \in \mathbb{R}^{(d \times K^{2}) \times h \times w}$ represent offsets and modulation scalars for the deformable convolution, respectively. Here, $d$ and $K^{2}$ denote the deformable group and the kernel size of the deformable convolution, respectively.
Then, modulated deformable convolution \cite{zhu2019deformable} with the predicted parameters $\Delta P$ and $\Delta M$ is applied to the feature $F_{i}^{c}$ to get the local fused feature $F_{i}^b$:
\begin{equation}
  F_{i}^{b}(p)=\sum_{k=1}^{ K^{2} } \omega_{k} \cdot F_{i}^{c}(p+p_{k}+\Delta p_{i,k}) \cdot \Delta m_{i,k},
\end{equation}
where $\omega_{k}$ represents the weights for each location $p$, and
$p_{k}$ denotes the sampling grid with $K^{2}$ sampling locations;
$\Delta p_{k}$ and $\Delta m_{k}$ are the learnable offset and modulation scalar for the $k$-th location, respectively. And $ \Delta P = \{\Delta p_{k}\}$, $ \Delta M = \{\Delta m_{k}\}$.
As the $p+p_{k}+\Delta p_{k}$ can be fractional, bilinear interpolation is adopted as in \cite{zhu2019deformable}.
In this way, 
both spatial dependence of intra-frame and local temporal dependence of inter-frame are modeled
by the designed LTMF module.

The structure of the AGRD block is shown in Fig. \ref{LTMF} (b). In
detail, 
to reduce computational costs  without compromising detection performance,
a $3 \times 3 $ convolutional layer is applied to reduce the number of channels by half.
Then,  since the spatial and temporal dependencies between  consecutive frames are 
divided into different channels, 
spatial attention and channel attention \cite{woo2018cbam} are applied
to model spatio-temporal dependency.
After that, the cascading  Residual Dense Blocks (RDBs) \cite{zhang2018residual} allow the network to have a large enough receptive field to further gather spatio-temporal information.
Finally, a $3 \times 3 $ convolutional layer is utilized to restore the number of channels.

Although DCN has the potential to capture motion cues of the targets, 
the training of DCN is very unstable and the overflow of offsets severely degrades the model performance \cite{chan2021understanding}.
Therefore, we introduce a Spatio-Temporal Fusion (STF) loss $\mathcal{L}_{STF}$ to improve the optimization of offsets: 
\begin{equation}
\mathcal{L}_{i,STF}^B=\mathcal{L}_1\left(F_i^E, F_i^b \right),
\label{STF-B}
\end{equation}
where $\mathcal{L}_{i,STF}^B$ denotes the backward STF loss of frame $I_i$, and $\mathcal{L}_1(\cdot,\cdot)$ is $L_1$ loss.

\subsubsection{Global temporal motion fusion module}
\begin{figure}[t]
  \centering
  \includegraphics[width=0.4\linewidth]{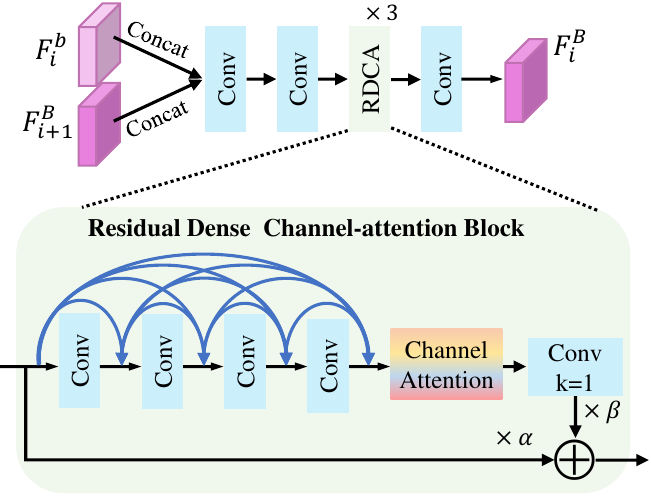}
  \caption{
  The structure of Global Temporal Motion Fusion (GTMF) module in backward propagation branch. 
  $F_{i+1}^B$ is the past propagation feature of the $i$-th frame,
which carries future global temporal information.
  }
  \label{GTMF}
\end{figure}

To utilize global distant frame information, 
as seen in Fig. \ref{GTMF},
we first use a $3 \times 3$ convolutional layer to fuse the propagation feature $F_{i+1}^B$ and the local fused feature $F_{i}^b$: 
\begin{equation}
F_{i}^m = Conv([F_{i+1}^B, F_{i}^b]), 
\end{equation}
where $F_{i}^m$ is the mixed feature.
Then, an intuitive way is using stacked convolutional layers to fuse concatenated feature.
However, when the input video clip is too long, 
the propagation feature is susceptible to error accumulation, 
especially when there is large motion between consecutive frames.
The directly concatenated feature may cause the generated target bounding boxes to be too large or even multiple additional targets to be detected (i.e., false detection).
Fortunately, the channel attention mechanism \cite{woo2018cbam} can allow the network to focus on more valuable channels and and improve discriminative learning capability.
Therefore, we use channel attention in RDB and propose Residual Dense Channel-attention (RDCA) block to alleviate the problem of error accumulation in the network.
We experimentally observed that inserting the channel-attention layer before dense fusion can achieve better performance.
Meanwhile, to  enhance the flexibility of merging  shallow and deep features, two additional learnable parameters $\alpha$ and $\beta$ are incorporated into the RDCA block, which are initialized with 1 and 0.2, respectively.
The new  propagation feature $F_{i}^B$ can be generated as follows:
\begin{equation}
F_{i}^B = Conv \circ   (RDCA)^3 \circ   Conv(F_{i}^m).
\end{equation}
The subsequent ablation study shows that the GTMF module can greatly improve detection performance without complex structural design.

{\bf Remark.} 
Although the existing learning-based multi-frame methods can increase the size of the sliding window to utilize information from more distant frames (i.e., global temporal information), 
they still have two problems: 1) 
They can only utilize the local temporal information of adjacent frames
within a narrow training sampling window, while ignoring the global
temporal information of distant frames outside the sliding
window. The essence is that once the models for these methods
are trained, the number of frames used in their inference and
training stages must remain consistent. 
2) Model parameters and computational cost will be increased. 
In contrast, our propagation scheme does not have the above problems.

\subsection{The loss function}
\label{3.3}
Following the common paradigm of  target detection, the detection loss $\mathcal{L}_{D}$ can be expressed as:
\begin{equation}
\mathcal{L}_{D}=\lambda \mathcal{L}_{r e g}+\mathcal{L}_{c l s}+\mathcal{L}_{o b j},
\label{d_loss}
\end{equation}
where $\mathcal{L}_{r e g}$ is the bounding box regression loss, $\mathcal{L}_{c l s}$ denotes the classification loss, and $\mathcal{L}_{o b j}$ refers to the target probability loss; 
$\lambda$ is a hyper-parameter. Following YOLOX \cite{ge2021yolox}, $\lambda$ is set to 5.

As described in Section \ref{sec3-1}, different from the existing methods, 
we generate detection results for one video clip at a time. 
Therefore, the loss function of our BIRD method can be represented as:
\begin{equation}
\mathcal{L}=\sum_{i=t}^{t+N} (\mathcal{L}_{i,D} + \eta \mathcal{L}_{i,STF}),
\end{equation}
where $ \eta$  is a trade-off hyper-parameter between the detection loss and 
STF loss, which is empirically set as 1; 
$\mathcal{L}_{i,D}$ refers to the detection loss of frame $I_i$;  $\mathcal{L}_{i, STF}$
denotes the bidirectional STF loss of frame $I_i$ and is expressed as:
\begin{equation}
\mathcal{L}_{i,STF}=\mathcal{L}_{i,STF}^B + \mathcal{L}_{i,STF}^F,
\end{equation}
where $\mathcal{L}_{i,STF}^B$ is computed by Eq. (\ref{STF-B}), and $\mathcal{L}_{i,STF}^F$ refers to the forward STF loss of frame $I_i$ and  is calculated as follows: 
\begin{equation}
\mathcal{L}_{i,STF}^F=\mathcal{L}_1\left(F_i^B, F_i^f \right).
\end{equation}

\section{Experiments}
\label{Sec4}
\subsection{Datasets and quantitative evaluation metrics}
\textit{Datasets.} Following SSTNet \cite{chen2024sstnet}, 
we conduct all experiments on two MISTD datasets, 
i.e., DAUB \cite{hui2019dataset} and IRDST \cite{sun2023receptive}.
For the DAUB dataset, there are 10  and 7 video sequences in the training and test sets, respectively.
For the IRDST dataset, there are 42  and 43 video sequences in the training and test sets, respectively.

\textit{Quantitative Evaluation Metrics.}
Following SSTNet \cite{chen2024sstnet},
we apply four quantitative evaluation metrics to evaluate the model performance, 
that is, 
mAP$_{50}$ (i.e., the mean average precision with the IoU threshold of 0.5), Precision (Pr), Recall (Re) and $F1$ score.
These  metrics are formulated as follows:
\begin{equation}
\text { Precision }=\frac{\mathrm{TP}}{\mathrm{TP}+\mathrm{FP}},
\end{equation}
\begin{equation}
\text { Recall }=\frac{\mathrm{TP}}{\mathrm{TP}+\mathrm{FN}},
\end{equation}
\begin{equation}
F1 = \frac{ 2 \times \text { Precision } \times \text { Recall } } { \text { Precision } + \text { Recall } },
\end{equation}
where TP, FP, and FN denote the number of true positives (correct detections), false positives (false detections), and false negatives (missing detections), respectively.

\subsection{Implementation details}
\textit{Network Settings.}
In the  feature extraction module, 
the kernel size of the convolutional layers is set to $3 \times 3$;
all convolutional layers have 48 filters (except for the last layer,  which
is set to 64). 
The deformable group and the kernel size of the deformable convolution
are set to 64  and $3 \times 3$, respectively.
The number of dense connection layers and the growth rate of channels in the  residual dense block are set to 4 and 32, respectively.
The other settings in  our BIRD method can be found in Section \ref{Sec3}.

\textit{Experimental Settings.}
In the training stage, following SSTNet \cite{chen2024sstnet}, 
the number of frames used is set to 5. 
We adopt the Adam optimizer with its default settings to train our model.
The training epoch and batch size are set to 20 and 2, respectively.
The learning rate is initially set to $2 \times 10^{-4}$ and retained throughout training.
In the inference stage, the number of frames used is set to 8. 
The resolution of input frames is reshaped into $544 \times 544$ in the training and inference stages.
All experiments are conducted on a NVIDIA GeForce RTX 4090D GPU.

\begin{table*}[htbp]
  \centering
    \scriptsize
  \caption{Overall performance comparison in terms of mAP$_{50}$ (\%), Precision (Pr) (\%), Recall (Re) (\%) and $F1$ (\%) score on the DAUB and IRDST datasets.
  }
    \setlength{\tabcolsep}{1.5mm}

    \begin{tabular}{c|c|cccc|cccc}
    \toprule
    \multirow{2}{*}{\textbf{Scheme }} & \multirow{2}{*}{\textbf{Methods }} & \multicolumn{4}{c|}{\textbf{DAUB}} & \multicolumn{4}{c}{\textbf{IRDST}} \\
\cline{3-10}
&   & \textbf{mAP$_{50}$} & \textbf{Pr} & \textbf{Re} & \textbf{F1} & \textbf{mAP$_{50}$} & \textbf{Pr} & \textbf{Re} & \textbf{F1} \\
    \hline
    \multicolumn{1}{c|}{\multirow{12}[2]{*}{\textbf{  \thead{Single-frame \\ based detection} }}} 
          & ACM    & 72.30 & 76.84 & 95.31 & 85.09 & 67.74 & 81.44 & 84.01 & 82.71 \\
          & UIUNet & 88.23 & 94.63 & 94.79 & 94.71 & 70.96 & 87.28 & 82.08 & 84.60 \\
          & ISNet   & 83.43 & 88.64 & 95.04 & 91.73 & 75.00 & 87.78 & 86.81 & 87.29 \\
          & AGPCNet   & 73.08 & 78.49 & 94.41 & 85.72 & 73.86 & 83.77 & 89.18 & 86.39 \\
          & SANet  & 87.90 & 94.14 & 94.22 & 94.18 & 77.98 & 85.42 & 92.13 & 88.64 \\
          & DNANet  & 89.24 & 95.66 & 94.83 & 95.24 & 76.84 & 90.08 & 86.81 & 88.42 \\
          & RDIAN   & 83.69 & 90.55 & 93.37 & 91.94 & 71.99 & 84.41 & 86.48 & 85.43 \\
          & SIRST5K  & 88.45 & 94.48 & 94.97 & 94.72 & 72.64 & 86.17 & 85.65 & 85.91 \\
          & RPCANet  & 85.75 & 89.12 & 97.58 & 93.16 & 73.29 & 85.02 & 87.13 & 86.06 \\
          & MSHNet  & 89.23 & 97.27 & 92.26 & 94.70 & 78.50 & 88.89 & 89.63 & 89.26 \\
          & SCTransNet   & 88.26 & 93.50 & 95.50 & 94.53 & 78.27 & 89.67 & 88.43 & 89.05 \\
    \midrule
    \multicolumn{1}{c|}{\multirow{3}[2]{*}{\textbf{ \thead{Multi-frame \\ based detection} }}} 
          & DTUM   & 88.24 & 95.15 & 93.60 & 94.37 & 80.98 & 90.62 & 90.46 & 90.54 \\
          & TMP   & 87.87 & 95.94 & 92.97 & 94.43 & 81.38 & 91.04 & 90.93 & 90.98 \\ 
          
          & SSTNet  &  {94.33} &  {97.77} &  {97.91} &  {97.84} &  {83.25} &  {91.13} &  {92.24} &  {91.68} \\

          & \textbf{BIRD (Ours)}   & \textbf{97.85} & \textbf{99.30} & \textbf{99.31} & \textbf{99.30} & \textbf{92.60} & \textbf{96.69} & \textbf{96.73} & \textbf{96.71} \\
    \bottomrule
    \end{tabular}%

  \label{Comparisons}%
\end{table*}%

\subsection{Comparisons with state-of-the-art methods}
We evaluate the effectiveness of the proposed BIRD approach by comparing it with many learning-based approaches, 
including 11 single-frame based methods: 
ACM \cite{dai2021asymmetric},  UIUNet \cite{wu2022uiu}, ISNet \cite{zhang2022isnet},   AGPCNet \cite{zhang2023attention}, SANet \cite{zhu2023sanet},  DNANet \cite{li2023dense}, RDIAN \cite{sun2023receptive}, SIRST5K \cite{10496142}, RPCANet \cite{wu2024rpcanet}, MSHNet \cite{liu2024infrared}  and SCTransNet \cite{10486932},
and 3 multi-frame based  methods: DTUM \cite{10321723}, TMP \cite{ZHU2024124731} and SSTNet \cite{chen2024sstnet}.
Since the two datasets we applied are based on bounding box annotations, 
and almost all single-frame detection methods (except the SANet method) and the DTUM method
are based on pixel-level segmentation.
To make the comparison as fair as possible,
following SSTNet \cite{chen2024sstnet},
we add the detection head YOLOX \cite{ge2021yolox} to the output of their networks to generate bounding boxes.
Then, we retrain these modified models and SANet, TMP and SSTNet methods  under the same experimental settings.

\begin{figure}[t]
  \centering
  \includegraphics[width=0.5\linewidth]{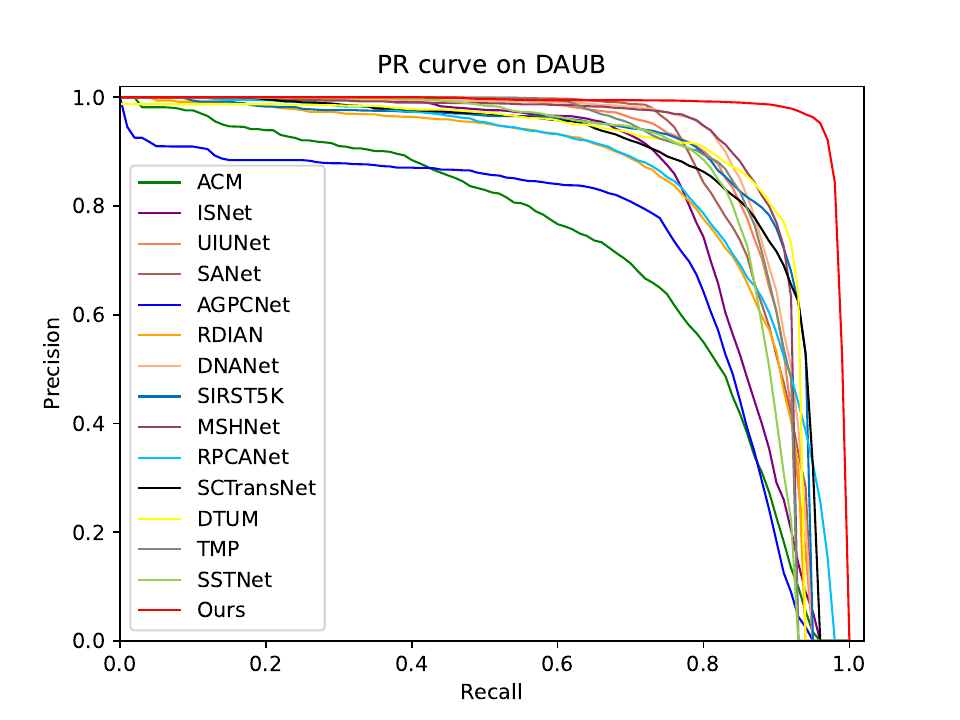}
  \caption{PR curve comparison on the DAUB dataset. 
  The area under the curve of our BIRD method (red curve) is larger than those of the other methods.
  }
  \label{p-r-DAUB}
\end{figure}

\subsubsection{Overall quantitative comparison}
Table \ref{Comparisons} shows the quantitative comparison results of different approaches on two datasets.
From Table \ref{Comparisons}, two apparent findings can be observed.

One is that our BIRD method achieves better performance than all comparison methods on two datasets in terms of four standard evaluation metrics. 
For example, on the DAUB dataset,
our method reaches the highest mAP$_{50}$ 97.85\%, 
 Pr 99.30\%, 
 Re 99.31\% and  $F1$ 99.30\%.
Furthermore, on the IRDST dataset, 
our method still reaches the highest mAP$_{50}$ 92.60\%, 
 Pr 96.69\%, 
 Re 96.73\% and  $F1$ 96.71\%, and significantly surpasses other compared methods.

The other is that our BIRD method has stronger data adaptability than
other compared methods.
Almost all  methods achieve superior performance
on the DAUB dataset in comparison to the IRDST dataset.
For example, 
the SSTNet method reaches mAP$_{50}$ 94.33\% and $F1$ 97.84\% on the DAUB dataset, 
but just reaches mAP$_{50}$ 83.25\% and $F1$ 91.68\% on the IRDST dataset.
The possible reason for this is that the IRDST dataset contains more complex scenes 
with the effects of cluttered background and noise.
Nevertheless, our method still performs well on the IRDST dataset.

\begin{figure}[t]
  \centering
  \includegraphics[width=0.5\linewidth]{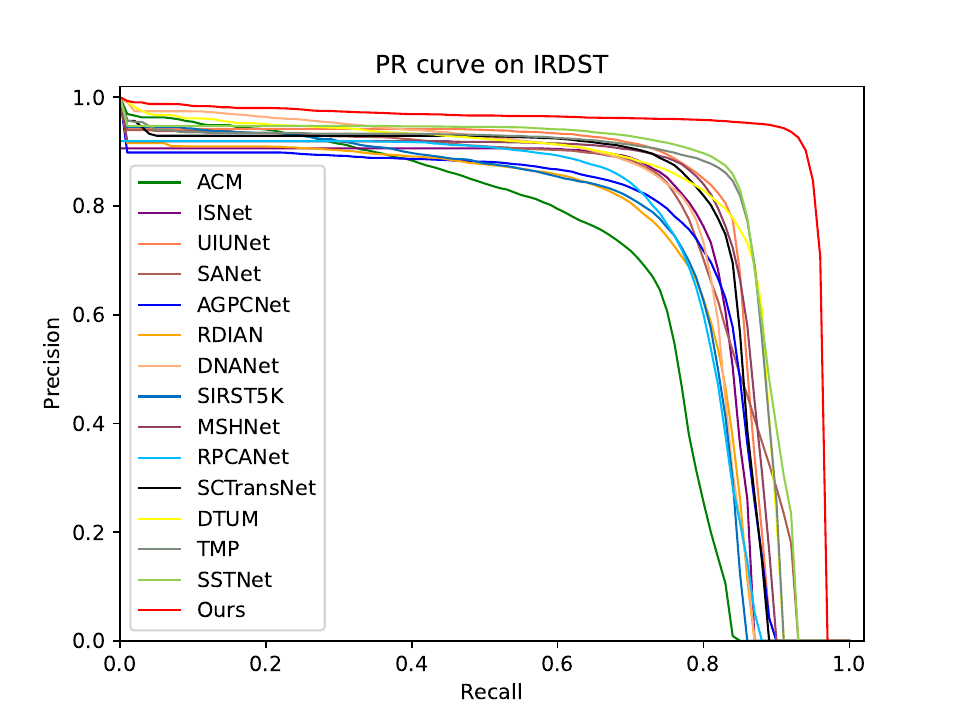}
  \caption{PR curve comparison on the IRDST dataset. The proposed BIRD method  distinctly outperforms other methods.}
  \label{p-r-IRDST}
\end{figure}

\subsubsection{Precision–recall (PR) curve comparison}
We use PR curve 
to further compare the comprehensive performance of different methods.
The comparison of PR curves on the DAUB and IRDST datasets is shown in Fig. \ref{p-r-DAUB}  and Fig. \ref{p-r-IRDST}, respectively.
It can be easily observed from Figs. \ref{p-r-DAUB} and \ref{p-r-IRDST} that our curves are almost always above those of other methods on two datasets.
In conclusion, 
comparison of the two groups of PR curves mean that
our BIRD method has the best overall performance and achieves the best balance between Precision and Recall.

\begin{table}[t]
  \centering
   \scriptsize
  \caption{
  Inference complexity comparison on the DAUB dataset.
  For a fair comparison, all methods are retested on a NVIDIA GeForce RTX 4090D.
  The results are reported by model Parameters (Params), Floating-point Operations (FLOPs) and Frame Per Second (FPS). 
  Frames indicates the number of frames used for training.}
    \setlength{\tabcolsep}{1.5mm}
  
    \begin{tabular}{l|c|cc|ccc}
    \toprule
    \textbf{Methods} &  \textbf{Frames} & \textbf{mAP$_{50}$} & \textbf{F1} & \textbf{Params (M)} & \textbf{FLOPs (G)} & \textbf{FPS}\\
    \midrule
    ACM   & 1 & 72.30 & 85.09 & 3.01  & \textbf{28.17} & 43.58  \\
    ISNet  & 1 & 83.39 & 91.90 & 3.48  & 300.64  & 17.13  \\
    UIUNet & 1 & 88.23 & 94.71 & 53.03  & 515.83  & 15.70  \\
    SANet  & 1 & 87.90 & 94.18 & 12.40  & 47.46  & 17.39  \\
    AGPCNet & 1 & 73.08 & 85.72 & 14.85  & 413.61  & 13.74  \\
    RDIAN & 1 & 83.69 & 91.94 & \textbf{2.71} & 57.37  & 36.97  \\
    DNANet & 1 & 89.24 & 95.24 & 7.19  & 152.58  & 9.23  \\
    SIRST5K  & 1 & 88.45 & 94.72 & 11.28  & 204.88  & 17.66  \\
    MSHNet  & 1 & 89.23 & 94.70 & 6.56  & 78.77  & 12.74  \\
    RPCANet & 1 & 85.75 & 93.16 & 3.18  & 432.27  & 14.75  \\
    SCTransNet  & 1 & 88.26 & 94.53 & 13.68  & 115.34  & 21.59  \\
    \midrule
    DTUM   & 5 & 88.24 & 94.37 & 2.79  & 117.10  & 21.23  \\
    TMP & 5 & 87.87 & 94.43 & 17.56  & 104.98 & 17.13  \\
    
    SSTNet  & 5 & 94.33 & 97.84 & 11.95  & 139.53  & 17.84  \\
    
    \textbf{Ours} & 5 & \textbf{97.85} & \textbf{99.30} & 9.39  & 98.36  & \textbf{55.08} \\
    \bottomrule
    \end{tabular}%
    
  \label{Speed_parameter}%
\end{table}%

\subsubsection{Model complexity comparison}
We use model parameters, FPS and FLOPs and to compare  model complexity of different methods.
As shown in Table \ref{Speed_parameter}, 
although RDIAN  method  has the smallest parameters of 2.71M and ACM method  has the smallest FLOPs of 28.17G, 
the proposed BIRD method 
achieves the fastest inference speed with a 
moderate number of parameters and FLOPs.
Specifically, since each frame in the video is processed only once, 
the inference speed of our method is more than three times faster than those of the compared multi-frame methods.
Meanwhile, our method still 
has lower FLOPs than the compared multi-frame methods.
It demonstrates the efficiency and effectiveness of the proposed method.

\begin{figure*}[t]
  \centering
  \includegraphics[width=0.85\linewidth]{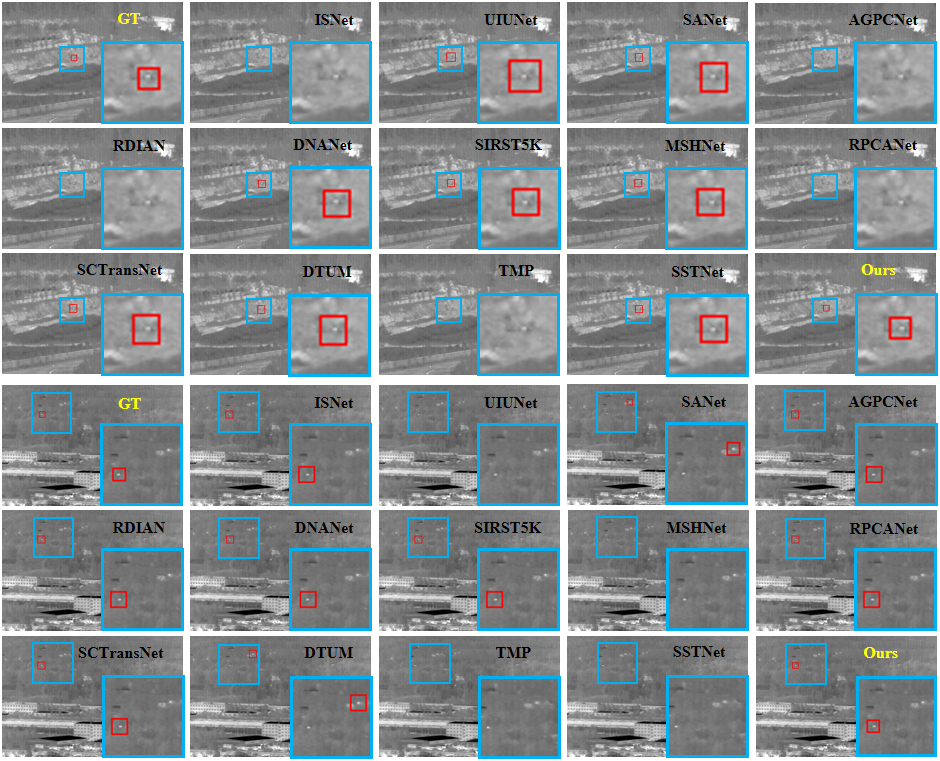}
  \caption{
Two groups of visualization comparisons on the DAUB dataset; GT: ground truth. 
Blue box  is the detection region and amplified;
red boxes are the generated target bounding boxes.
}
  \label{Visualization_DAUB}
\end{figure*}

\subsubsection{Visualization comparison}
\begin{figure*}[t]
  \centering
  \includegraphics[width=0.85\linewidth]{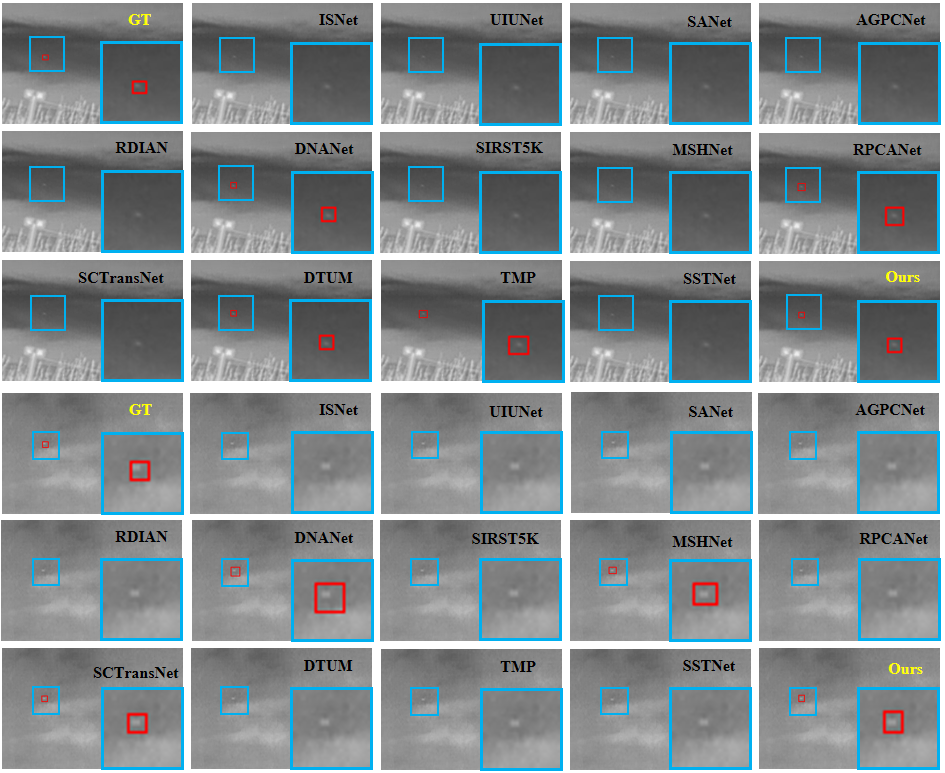}
  \caption{
Two groups of visualization comparisons on the IRDST dataset; GT: ground truth. 
Blue box  is the detection region and amplified;
red boxes are the generated target bounding boxes.
}
  \label{Visualization_IRDST}
\end{figure*}

We present two groups of visual detection results of different methods on the DAUB
and IRDST datasets for qualitative comparison in Fig. \ref{Visualization_DAUB} and Fig. \ref{Visualization_IRDST}, respectively.
From Figs. \ref{Visualization_DAUB} and \ref{Visualization_IRDST}
We can see 
that our approach can usually precisely detect
targets. In contrast, other approaches often
lead to false detections or missed detections.

For example, on the DAUB dataset, in the first group of  comparison (the top three rows), our approach can  accurately  detect the target, while ISNet, AGPCNet, RDIAN, RPCANet and TMP approaches occur missed detections.
Although UIUNet and SANet approaches can detect target, the generated target bounding boxes appear too large and imprecise.
Moreover, in the second group of visualization comparison (the bottom three rows), 
UIUNet, MSHNet, TMP and SSTNet approaches cause missed detections; 
SANet and  DTUM approaches produce false detections.

In addition, on the IRDST dataset,
almost all approaches produce missed detections since the two groups of targets are hard to perceive, while our approach can still detect the targets precisely. 
One possible reason is that our bidirectional propagation branch can utilize temporal information from more distant frames.

In a word, the qualitative results of these visualization comparisons 
are highly consistent with  the quantitative results in Table \ref{Comparisons},
demonstrating the
advantage of our BIRD method.

\begin{table*}[t]
  \centering
  \scriptsize
  \caption{Ablation study of the proposed Backward Propagation (BP) branch, Forward
 Propagation (FP) branch and Joint Optimization Strategy (JOS).} 
    \setlength{\tabcolsep}{2mm}
    
    \begin{tabular}{ccc|cccc|cccc|c}
    \toprule
    \multirow{2}[4]{*}{\textbf{BP}} & \multirow{2}[4]{*}{\textbf{FP}} & \multirow{2}[4]{*}{\textbf{JOS}} & \multicolumn{4}{c|}{\textbf{DAUB}} & \multicolumn{4}{c|}{\textbf{IRDST}} & \multirow{2}[4]{*}{\textbf{FPS}} \\
\cmidrule{4-11}          &       &       & \textbf{mAP$_{50}$} & \textbf{Pr} & \textbf{Re} & \textbf{F1} & \textbf{mAP$_{50}$} & \textbf{Pr} & \textbf{Re} & \textbf{F1} &  \\
    \midrule
    -     & -     & -     & 86.02  & 90.04  & 96.89  & 93.34  & 75.10  & 87.15  & 89.71  & 88.41  & 21.89  \\
    -     & -     & $\checkmark$     & 87.69  & 90.12  & 98.05  & 93.92  & 78.96  & 88.56  & 91.22  & 89.87  & \textbf{82.08} \\
    $\checkmark$     & -     & $\checkmark$     & 93.13  & 95.09  & 98.52  & 96.77  & 86.09  & 92.21  & 94.48  & 93.33  & 65.12  \\
    -     & $\checkmark$     & $\checkmark$     & 94.00  & 97.19  & 97.64  & 97.41  & 86.56  & 93.12  & 94.35  & 93.73  & 65.50  \\
    $\checkmark$     & $\checkmark$     & $\checkmark$     & \textbf{97.85} & \textbf{99.30} & \textbf{99.31} & \textbf{99.30} & \textbf{92.60} & \textbf{96.69} & \textbf{96.73} & \textbf{96.71} & 55.08  \\
    \bottomrule
    \end{tabular}%

  \label{Ablation}%
\end{table*}%

\begin{table}[htbp]
  \centering
   \scriptsize
  \caption{Ablation study on propagation branch.}
    \setlength{\tabcolsep}{1.2mm}

    \begin{tabular}{ccc|cc|cc}
    \toprule
    \multirow{2}[4]{*}{\textbf{LTMF}} & \multirow{2}[4]{*}{\textbf{STF }} & \multirow{2}[4]{*}{\textbf{GTMF}} & \multicolumn{2}{c|}{\textbf{DAUB}} & \multicolumn{2}{c}{\textbf{IRDST}} \\
\cmidrule{4-7}          &       &       & \textbf{mAP$_{50}$ (\%)} & \textbf{F1 (\%)} & \textbf{mAP$_{50}$ (\%)} & \textbf{F1 (\%)} \\
    \midrule
    -     & -     & -     & 86.02  & 93.34  & 75.10  & 88.41  \\
    $\checkmark$     & -     & -     & 91.80  & 96.16  & 85.99  & 93.17  \\
    $\checkmark$     & $\checkmark$     & -     & 92.40  & 96.50  & 87.70  & 94.23  \\
    -     & -     & $\checkmark$     & 92.51  & 96.77  & 86.25  & 93.56  \\
    $\checkmark$     & -     & $\checkmark$     &   96.64    &   98.73    &  90.91     & 95.84 \\
    $\checkmark$     & $\checkmark$     & $\checkmark$     & \textbf{97.85} & \textbf{99.30} & \textbf{92.60} & \textbf{96.71} \\
    \bottomrule
    \end{tabular}%

  \label{Ablation_propagation}%
\end{table}%

\subsection{Ablation study}
\subsubsection{Effects of different assemblies}
\label{Sec5}
To evaluate the effectiveness of our bidirectional propagation branches and 
joint optimization strategy (JOS),
we set up a baseline model.
In detail, we still use 5 frames as input but only  output the detection result of  the middle frame at once.
Specifically, we directly utilize a $3 \times 3$ convolutional layer with 192 filters to fuse consecutive three  frames (i.e., local temporal information), and then input the fused feature into the detection head.
Then,  different assemblies are inserted into the baseline, and these models are retrained with the identical experimental settings.
The experimental results are presented in Table \ref{Ablation}.

\textit{Effectiveness of Joint Optimization Strategy.}
We output the detection results of multiple frames at once to evaluate the effectiveness of the JOS.
From the results in Table \ref{Ablation}, 
we can see that although JOS does not significantly improve the performance, it can greatly increase the inference speed, 
which benefits from the fact that each frame only needs to be processed once.

\textit{Effectiveness of Bidirectional Propagation Branch.}
From the results in Table \ref{Ablation}, 
we can conclude that the unidirectional propagation scheme contributes a lot to
performance improvement, and the bidirectional propagation scheme is optimal.
For a thorough  investigation, we conduct an additional ablation study about the propagation branch configuration.
In detail, we use the
LTMF module to replace the simple local fusion operation
(i.e., concatenation operation) to evaluate its effectiveness.
We insert STF loss $\mathcal{L}_{STF}$ and GTMF module into the different 
models to evaluate their effectiveness.
The experimental results are presented in Table \ref{Ablation_propagation}.
The results in the 1st and 2nd
rows (or 4th and 5th rows) prove the effectiveness
of our proposed LTMF module.
Meanwhile,  the results in Table \ref{Ablation_propagation}
demonstrate that the GTMF module can further improve performance by utilizing global distant frame information.
In addition, 
from the results in the 5th and 6th rows, we can see that without
the supervision of STF loss, the performance
is still sub-optimal.
The reason is that deformable
convolution is intrinsically difficult to train, and training instability tends to leads to offset overflow \cite{chan2021understanding}, which degrades the final
performance. 
Therefore, it is crucial to use an additional loss for deformable fusion.

\begin{figure}[t]
  \centering
  \includegraphics[width=0.5\linewidth]{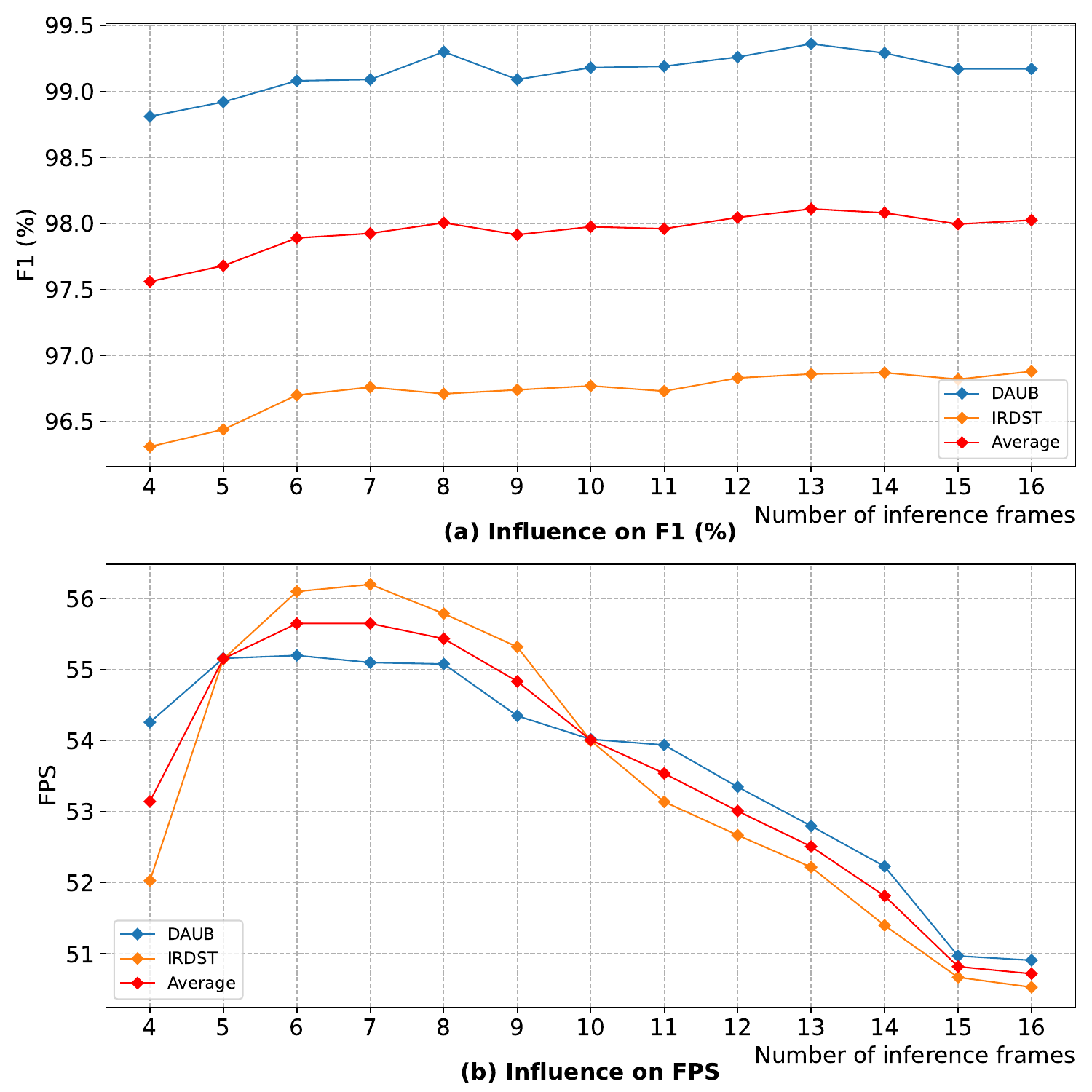}
  \caption{The influence of the number of inference frames on two datasets.}
  \label{F1_FPS}
\end{figure}

\subsubsection{Influence of the number of inference frames} 
As described in Section \ref{3.2}, 
theoretically, once our model is trained, 
our model can use any number of frames as input in the inference stage, 
so an ablation study is conducted to evaluate the impact of the number of inference frames.
To preserve the essence of our algorithm (i.e., utilizing both local and global temporal information), 
we start the study with a frame count of 4.
 As shown in Fig. \ref{F1_FPS}, 
 the inference speed first increases and then decreases as the number of inference frames increases.
 Meanwhile, the $F1$ score first increases and then gradually stabilizes as the number of inference frames increases.
 In addition, since our algorithm needs to utilize future
frame information, 
real-time processing latency is also reduced if fewer input frames are used in the inference stage.
Therefore, after comprehensively considering performance, inference speed and latency time, we set the number of inference frames to 8.

\begin{figure}[t]
  \centering
  \includegraphics[width=0.5\linewidth]{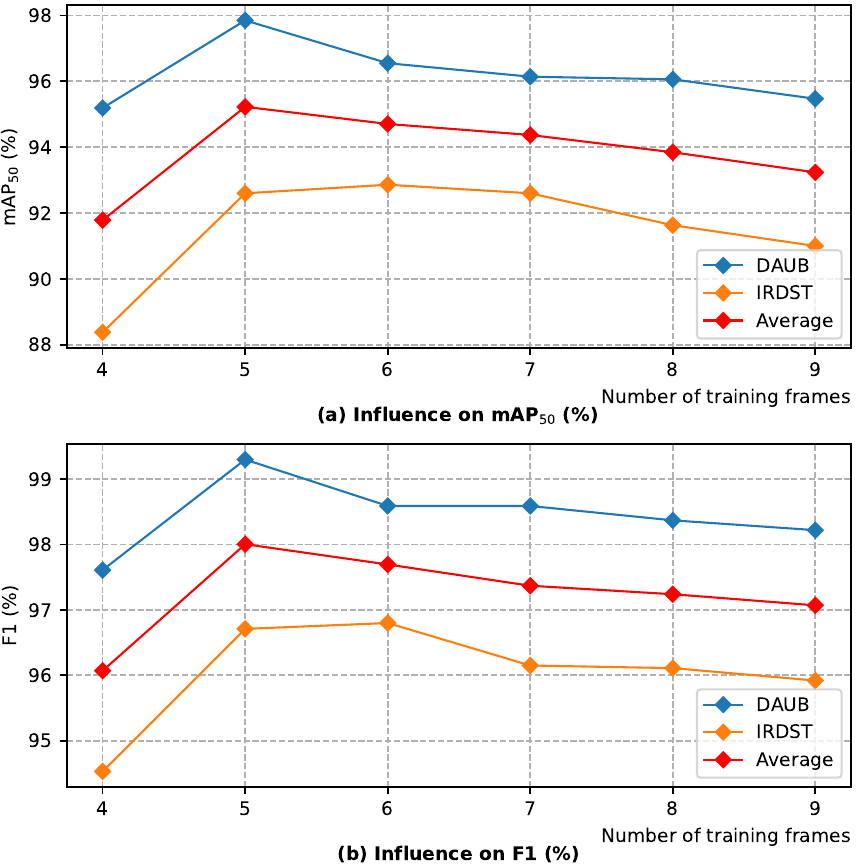}
  \caption{The impact of the number of training frames on two datasets.}
  \label{MAP50_F1}
\end{figure}

\subsubsection{Impact of the number of training frames} 
Even though 
we follow SSTNet approach \cite{chen2024sstnet} to use 5 frames as input in the training stage,
we also perform an ablation study to analyze the impact of the number of training frames on performance.
Similarly, we also start the study with a frame count of 4, 
and use 8 frames as input in the inference stage.
Due to  the constraints of our memory, the maximum number of training frame is set to 9.
We can see from Fig. \ref{MAP50_F1} that for the DAUB dataset, the optimal number of training frames is 5, and for the IRDST dataset, the optimal number of training frames is 6.
After comprehensively considering the performance on two datasets, we also use 5 frames for model training.

\section{Conclusion}
In this paper, we proposed a BIRD method to address MISTD task in a new fashion (i.e., recursive fashion).
Specifically, a LTMF module is designed based on deformable convolution 
to align and utilize local temporal information. Meanwhile, we developed a GTMF module to aggregate global bidirectional temporal information.
A spatio-temporal fusion loss is further introduced to improve the alignment and fusion of local temporal information. Extensive experiments are conducted to show the efficiency and effectiveness of the proposed BIRD method. Ablation studies are also conducted to show the impacts of different assemblies and the flexibility of our method.




%




\bibliographystyle{IEEEtran}
\bibliography{References}

\begin{thebibliography}{10}
\providecommand{\url}[1]{#1}
\csname url@samestyle\endcsname
\providecommand{\newblock}{\relax}
\providecommand{\bibinfo}[2]{#2}
\providecommand{\BIBentrySTDinterwordspacing}{\spaceskip=0pt\relax}
\providecommand{\BIBentryALTinterwordstretchfactor}{4}
\providecommand{\BIBentryALTinterwordspacing}{\spaceskip=\fontdimen2\font plus
\BIBentryALTinterwordstretchfactor\fontdimen3\font minus \fontdimen4\font\relax}
\providecommand{\BIBforeignlanguage}[2]{{%
\expandafter\ifx\csname l@#1\endcsname\relax
\typeout{** WARNING: IEEEtran.bst: No hyphenation pattern has been}%
\typeout{** loaded for the language `#1'. Using the pattern for}%
\typeout{** the default language instead.}%
\else
\language=\csname l@#1\endcsname
\fi
#2}}
\providecommand{\BIBdecl}{\relax}
\BIBdecl

\bibitem{peng2023courtnet}
J.~Peng, H.~Zhao, K.~Zhao, Z.~Wang, and L.~Yao, ``Courtnet: Dynamically balance the precision and recall rates in infrared small target detection,'' \emph{Expert Systems with Applications}, vol. 233, p. 120996, 2023.

\bibitem{xu2021video}
C.~Xu, Z.~Gao, H.~Zhang, S.~Li, and V.~H.~C. de~Albuquerque, ``Video salient object detection using dual-stream spatiotemporal attention,'' \emph{Applied Soft Computing}, vol. 108, p. 107433, 2021.

\bibitem{9747993}
L.~Han and Z.~Yin, ``Global memory and local continuity for video object detection,'' \emph{IEEE Transactions on Multimedia}, vol.~25, pp. 3681--3693, 2023.

\bibitem{10173615}
Q.~Qi, Y.~Yan, and H.~Wang, ``Class-aware dual-supervised aggregation network for video object detection,'' \emph{IEEE Transactions on Multimedia}, vol.~26, pp. 2109--2123, 2024.

\bibitem{wang2024tiny}
G.~Wang, X.~Yang, L.~Li, K.~Gao, J.~Gao, J.-y. Zhang, D.-j. Xing, and Y.-z. Wang, ``Tiny drone object detection in videos guided by the bio-inspired magnocellular computation model,'' \emph{Applied Soft Computing}, p. 111892, 2024.

\bibitem{chen2024dila}
P.~Chen, J.~Wang, Z.~Zhang, and C.~He, ``Dila: Dynamic gaussian distribution fitting and imitation learning-based label assignment for tiny object detection,'' \emph{Applied Soft Computing}, vol. 164, p. 111980, 2024.

\bibitem{10298041}
F.~Liu, C.~Gao, F.~Chen, D.~Meng, W.~Zuo, and X.~Gao, ``Infrared small and dim target detection with transformer under complex backgrounds,'' \emph{IEEE Transactions on Image Processing}, vol.~32, pp. 5921--5932, 2023.

\bibitem{feng2024meta}
H.~Feng, W.~Tang, H.~Xu, C.~Jiang, S.~S. Ge, and J.~He, ``Meta-learning based infrared ship object detection model for generalization to unknown domains,'' \emph{Applied Soft Computing}, vol. 159, p. 111633, 2024.

\bibitem{kwan2020enhancing}
C.~Kwan and B.~Budavari, ``Enhancing small moving target detection performance in low-quality and long-range infrared videos using optical flow techniques,'' \emph{Remote Sensing}, vol.~12, no.~24, p. 4024, 2020.

\bibitem{zhao2019infrared}
F.~Zhao, T.~Wang, S.~Shao, E.~Zhang, and G.~Lin, ``Infrared moving small-target detection via spatiotemporal consistency of trajectory points,'' \emph{IEEE Geoscience and Remote Sensing Letters}, vol.~17, no.~1, pp. 122--126, 2020.

\bibitem{sun2019infrared}
Y.~Sun, J.~Yang, Y.~Long, and W.~An, ``Infrared small target detection via spatial-temporal total variation regularization and weighted tensor nuclear norm,'' \emph{IEEE Access}, vol.~7, pp. 56\,667--56\,682, 2019.

\bibitem{wu2023infrared}
F.~Wu, H.~Yu, A.~Liu, J.~Luo, and Z.~Peng, ``Infrared small target detection using spatiotemporal 4-d tensor train and ring unfolding,'' \emph{IEEE Transactions on Geoscience and Remote Sensing}, vol.~61, pp. 1--22, 2023.

\bibitem{chen2024tci}
T.~Chen, Z.~Tan, Q.~Chu, Y.~Wu, B.~Liu, and N.~Yu, ``Tci-former: Thermal conduction-inspired transformer for infrared small target detection,'' in \emph{Proceedings of the AAAI Conference on Artificial Intelligence}, 2024, pp. 1201--1209.

\bibitem{LIN2024124385}
J.~Lin, S.~Li, X.~Yang, S.~Niu, B.~Yan, and Z.~Meng, ``Cs-vig-unet: Infrared small and dim target detection based on cycle shift vision graph convolution network,'' \emph{Expert Systems with Applications}, vol. 254, p. 124385, 2024.

\bibitem{chen2024unveiling}
H.~Chen, X.~Sun, C.~Hu, H.~Wang, and J.~Peng, ``Unveiling the power of haar frequency domain: Advancing small target motion detection in dim light,'' \emph{Applied Soft Computing}, vol. 167, p. 112281, 2024.

\bibitem{du2021multiple}
J.~Du, D.~Li, Y.~Deng, L.~Zhang, H.~Lu, M.~Hu, X.~Shen, Z.~Liu, and X.~Ji, ``Multiple frames based infrared small target detection method using cnn,'' in \emph{Proceedings of the 2021 4th International Conference on Algorithms, Computing and Artificial Intelligence}, 2021, pp. 1--6.

\bibitem{du2021spatial}
J.~Du, H.~Lu, L.~Zhang, M.~Hu, S.~Chen, Y.~Deng, X.~Shen, and Y.~Zhang, ``A spatial-temporal feature-based detection framework for infrared dim small target,'' \emph{IEEE Transactions on Geoscience and Remote Sensing}, vol.~60, pp. 1--12, 2021.

\bibitem{yan2023stdmanet}
P.~Yan, R.~Hou, X.~Duan, C.~Yue, X.~Wang, and X.~Cao, ``Stdmanet: Spatio-temporal differential multiscale attention network for small moving infrared target detection,'' \emph{IEEE Transactions on Geoscience and Remote Sensing}, vol.~61, pp. 1--16, 2023.

\bibitem{10521471}
H.~Deng, Y.~Zhang, Y.~Li, K.~Cheng, and Z.~Chen, ``Bemst: Multiframe infrared small-dim target detection using probabilistic estimation of sequential backgrounds,'' \emph{IEEE Transactions on Geoscience and Remote Sensing}, vol.~62, pp. 1--15, 2024.

\bibitem{ZHU2024124731}
S.~Zhu, L.~Ji, J.~Zhu, S.~Chen, and W.~Duan, ``Tmp: Temporal motion perception with spatial auxiliary enhancement for moving infrared dim-small target detection,'' \emph{Expert Systems with Applications}, vol. 255, p. 124731, 2024.

\bibitem{10275009}
Z.~Zhang, P.~Gao, S.~Ji, X.~Wang, and P.~Zhang, ``Infrared small target detection combining deep spatial–temporal prior with traditional priors,'' \emph{IEEE Transactions on Geoscience and Remote Sensing}, vol.~61, pp. 1--18, 2023.

\bibitem{10321723}
R.~Li, W.~An, C.~Xiao, B.~Li, Y.~Wang, M.~Li, and Y.~Guo, ``Direction-coded temporal u-shape module for multiframe infrared small target detection,'' \emph{IEEE Transactions on Neural Networks and Learning Systems}, pp. 1--14, 2023.

\bibitem{tong2024st}
X.~Tong, Z.~Zuo, S.~Su, J.~Wei, X.~Sun, P.~Wu, and Z.~Zhao, ``St-trans: Spatial-temporal transformer for infrared small target detection in sequential images,'' \emph{IEEE Transactions on Geoscience and Remote Sensing}, vol.~62, pp. 1--19, 2024.

\bibitem{liu2021dim}
X.~Liu, X.~Li, L.~Li, X.~Su, and F.~Chen, ``Dim and small target detection in multi-frame sequence using bi-conv-lstm and 3d-conv structure,'' \emph{IEEE Access}, vol.~9, pp. 135\,845--135\,855, 2021.

\bibitem{chen2024sstnet}
S.~Chen, L.~Ji, J.~Zhu, M.~Ye, and X.~Yao, ``{SSTNet}: Sliced spatio-temporal network with cross-slice convlstm for moving infrared dim-small target detection,'' \emph{IEEE Transactions on Geoscience and Remote Sensing}, vol.~62, pp. 1--12, 2024.

\bibitem{zhu2020tnlrs}
H.~Zhu, H.~Ni, S.~Liu, G.~Xu, and L.~Deng, ``Tnlrs: Target-aware non-local low-rank modeling with saliency filtering regularization for infrared small target detection,'' \emph{IEEE Transactions on Image Processing}, vol.~29, pp. 9546--9558, 2020.

\bibitem{gao2013infrared}
C.~Gao, D.~Meng, Y.~Yang, Y.~Wang, X.~Zhou, and A.~G. Hauptmann, ``Infrared patch-image model for small target detection in a single image,'' \emph{IEEE Transactions on Image Processing}, vol.~22, no.~12, pp. 4996--5009, 2013.

\bibitem{liu2023combining}
T.~Liu, Q.~Yin, J.~Yang, Y.~Wang, and W.~An, ``Combining deep denoiser and low-rank priors for infrared small target detection,'' \emph{Pattern Recognition}, vol. 135, p. 109184, 2023.

\bibitem{deshpande1999max}
S.~D. Deshpande, M.~H. Er, R.~Venkateswarlu, and P.~Chan, ``Max-mean and max-median filters for detection of small targets,'' in \emph{Signal and Data Processing of Small Targets 1999}, 1999, pp. 74--83.

\bibitem{bai2010analysis}
X.~Bai and F.~Zhou, ``Analysis of new top-hat transformation and the application for infrared dim small target detection,'' \emph{Pattern Recognition}, vol.~43, no.~6, pp. 2145--2156, 2010.

\bibitem{chen2013local}
C.~P. Chen, H.~Li, Y.~Wei, T.~Xia, and Y.~Y. Tang, ``A local contrast method for small infrared target detection,'' \emph{IEEE Transactions on Geoscience and Remote Sensing}, vol.~52, no.~1, pp. 574--581, 2013.

\bibitem{moradi2020fast}
S.~Moradi, P.~Moallem, and M.~F. Sabahi, ``Fast and robust small infrared target detection using absolute directional mean difference algorithm,'' \emph{Signal Processing}, vol. 177, p. 107727, 2020.

\bibitem{wang2019miss}
H.~Wang, L.~Zhou, and L.~Wang, ``Miss detection vs. false alarm: Adversarial learning for small object segmentation in infrared images,'' in \emph{Proceedings of the IEEE/CVF International Conference on Computer Vision}, 2019, pp. 8509--8518.

\bibitem{dai2021asymmetric}
Y.~Dai, Y.~Wu, F.~Zhou, and K.~Barnard, ``Asymmetric contextual modulation for infrared small target detection,'' in \emph{Proceedings of the IEEE/CVF Winter Conference on Applications of Computer Vision}, 2021, pp. 950--959.

\bibitem{wang2022interior}
K.~Wang, S.~Du, C.~Liu, and Z.~Cao, ``Interior attention-aware network for infrared small target detection,'' \emph{IEEE Transactions on Geoscience and Remote Sensing}, vol.~60, pp. 1--13, 2022.

\bibitem{zhang2022rkformer}
M.~Zhang, H.~Bai, J.~Zhang, R.~Zhang, C.~Wang, J.~Guo, and X.~Gao, ``Rkformer: Runge-kutta transformer with random-connection attention for infrared small target detection,'' in \emph{Proceedings of the 30th ACM International Conference on Multimedia}, 2022, pp. 1730--1738.

\bibitem{10295542}
J.~Lin, S.~Li, L.~Zhang, X.~Yang, B.~Yan, and Z.~Meng, ``Ir-transdet: Infrared dim and small target detection with ir-transformer,'' \emph{IEEE Transactions on Geoscience and Remote Sensing}, vol.~61, pp. 1--13, 2023.

\bibitem{sun2023receptive}
H.~Sun, J.~Bai, F.~Yang, and X.~Bai, ``Receptive-field and direction induced attention network for infrared dim small target detection with a large-scale dataset irdst,'' \emph{IEEE Transactions on Geoscience and Remote Sensing}, vol.~61, pp. 1--13, 2023.

\bibitem{10288394}
F.~Lin, S.~Ge, K.~Bao, C.~Yan, and D.~Zeng, ``Learning shape-biased representations for infrared small target detection,'' \emph{IEEE Transactions on Multimedia}, vol.~26, pp. 4681--4692, 2024.

\bibitem{zhang2005detecting}
F.~Zhang, C.~Li, and L.~Shi, ``Detecting and tracking dim moving point target in ir image sequence,'' \emph{Infrared Physics \& Technology}, vol.~46, no.~4, pp. 323--328, 2005.

\bibitem{ren2019infrared}
X.~Ren, J.~Wang, T.~Ma, K.~Bai, M.~Ge, and Y.~Wang, ``Infrared dim and small target detection based on three-dimensional collaborative filtering and spatial inversion modeling,'' \emph{Infrared physics \& technology}, vol. 101, pp. 13--24, 2019.

\bibitem{LUO2026111894}
D.~Luo, Y.~Xiang, H.~Wang, L.~Ji, S.~Li, and M.~Ye, ``Deformable feature alignment and refinement for moving infrared small target detection,'' \emph{Pattern Recognition}, p. 111894, 2025.

\bibitem{ge2021yolox}
Z.~Ge, S.~Liu, F.~Wang, Z.~Li, and J.~Sun, ``Yolox: Exceeding yolo series in 2021,'' \emph{arXiv preprint arXiv:2107.08430}, 2021.

\bibitem{dai2017deformable}
J.~Dai, H.~Qi, Y.~Xiong, Y.~Li, G.~Zhang, H.~Hu, and Y.~Wei, ``Deformable convolutional networks,'' in \emph{Proceedings of the IEEE International Conference on Computer Vision}, 2017, pp. 764--773.

\bibitem{9328265}
Z.~Shi, X.~Liu, K.~Shi, L.~Dai, and J.~Chen, ``Video frame interpolation via generalized deformable convolution,'' \emph{IEEE Transactions on Multimedia}, vol.~24, pp. 426--439, 2022.

\bibitem{luo2023spatio}
D.~Luo, M.~Ye, S.~Li, C.~Zhu, and X.~Li, ``Spatio-temporal detail information retrieval for compressed video quality enhancement,'' \emph{IEEE Transactions on Multimedia}, vol.~25, pp. 6808--6820, 2023.

\bibitem{zhu2019deformable}
X.~Zhu, H.~Hu, S.~Lin, and J.~Dai, ``Deformable convnets v2: More deformable, better results,'' in \emph{Proceedings of the IEEE/CVF conference on computer vision and pattern recognition}, 2019, pp. 9308--9316.

\bibitem{woo2018cbam}
S.~Woo, J.~Park, J.-Y. Lee, and I.~S. Kweon, ``Cbam: Convolutional block attention module,'' in \emph{Proceedings of the European Conference on Computer Vision (ECCV)}, 2018, pp. 3--19.

\bibitem{zhang2018residual}
Y.~Zhang, Y.~Tian, Y.~Kong, B.~Zhong, and Y.~Fu, ``Residual dense network for image super-resolution,'' in \emph{Proceedings of the IEEE Conference on Computer Vision and Pattern Recognition}, 2018, pp. 2472--2481.

\bibitem{chan2021understanding}
K.~C. Chan, X.~Wang, K.~Yu, C.~Dong, and C.~C. Loy, ``Understanding deformable alignment in video super-resolution,'' in \emph{Proceedings of the AAAI conference on artificial intelligence}, 2021, pp. 973--981.

\bibitem{hui2019dataset}
B.~Hui, Z.~Song, H.~Fan, P.~Zhong, W.~Hu, X.~Zhang, J.~Lin, H.~Su, W.~Jin, Y.~Zhang \emph{et~al.}, ``A dataset for infrared image dim-small aircraft target detection and tracking under ground/air background,'' \emph{Sci. Data Bank}, vol.~5, no.~12, p.~4, 2019.

\bibitem{wu2022uiu}
X.~Wu, D.~Hong, and J.~Chanussot, ``{UIU-Net}: U-net in u-net for infrared small object detection,'' \emph{IEEE Transactions on Image Processing}, vol.~32, pp. 364--376, 2022.

\bibitem{zhang2022isnet}
M.~Zhang, R.~Zhang, Y.~Yang, H.~Bai, J.~Zhang, and J.~Guo, ``{ISNet}: Shape matters for infrared small target detection,'' in \emph{Proceedings of the IEEE/CVF Conference on Computer Vision and Pattern Recognition}, 2022, pp. 877--886.

\bibitem{zhang2023attention}
T.~Zhang, L.~Li, S.~Cao, T.~Pu, and Z.~Peng, ``Attention-guided pyramid context networks for detecting infrared small target under complex background,'' \emph{IEEE Transactions on Aerospace and Electronic Systems}, vol.~59, no.~4, pp. 4250--4261, 2023.

\bibitem{zhu2023sanet}
J.~Zhu, S.~Chen, L.~Li, and L.~Ji, ``Sanet: Spatial attention network with global average contrast learning for infrared small target detection,'' in \emph{ICASSP 2023-2023 IEEE International Conference on Acoustics, Speech and Signal Processing (ICASSP)}, 2023, pp. 1--5.

\bibitem{li2023dense}
B.~Li, C.~Xiao, L.~Wang, Y.~Wang, Z.~Lin, M.~Li, W.~An, and Y.~Guo, ``Dense nested attention network for infrared small target detection,'' \emph{IEEE Transactions on Image Processing}, vol.~32, pp. 1745--1758, 2023.

\bibitem{10496142}
Y.~Lu, Y.~Lin, H.~Wu, X.~Xian, Y.~Shi, and L.~Lin, ``Sirst-5k: Exploring massive negatives synthesis with self-supervised learning for robust infrared small target detection,'' \emph{IEEE Transactions on Geoscience and Remote Sensing}, vol.~62, pp. 1--11, 2024.

\bibitem{wu2024rpcanet}
F.~Wu, T.~Zhang, L.~Li, Y.~Huang, and Z.~Peng, ``Rpcanet: Deep unfolding rpca based infrared small target detection,'' in \emph{Proceedings of the IEEE/CVF Winter Conference on Applications of Computer Vision}, 2024, pp. 4809--4818.

\bibitem{liu2024infrared}
Q.~Liu, R.~Liu, B.~Zheng, H.~Wang, and Y.~Fu, ``Infrared small target detection with scale and location sensitivity,'' in \emph{Proceedings of the IEEE/CVF Conference on Computer Vision and Pattern Recognition}, 2024, pp. 17\,490--17\,499.

\bibitem{10486932}
S.~Yuan, H.~Qin, X.~Yan, N.~Akhtar, and A.~Mian, ``Sctransnet: Spatial-channel cross transformer network for infrared small target detection,'' \emph{IEEE Transactions on Geoscience and Remote Sensing}, vol.~62, pp. 1--15, 2024.

\end{thebibliography}

\newpage

\vfill

\end{document}